\documentclass[10pt,twocolumn,letterpaper]{article}
\pdfoutput=1

\usepackage{iccv}
\usepackage{times}
\usepackage{epsfig}
\usepackage{graphicx}
\usepackage{amsmath}
\usepackage{amssymb}
\usepackage{multirow}
\usepackage{bbding}
\usepackage{color}
\usepackage{algorithm}
\usepackage{algorithmic}

\definecolor{orange}{rgb}{1,0.5,0}

\usepackage[breaklinks=true,bookmarks=false]{hyperref}

\iccvfinalcopy 

\def\iccvPaperID{3050} 

\begin{document}

\title{VoxelFormer: Bird’s-Eye-View Feature Generation based on Dual-view \\ Attention for Multi-view 3D Object Detection}

\author{Zhuoling Li$^{1}$\thanks{Zhuoling Li and Chuanrui Zhang contribute equally.}, \quad Chuanrui Zhang$^{1*}$, \quad Wei-Chiu Ma$^{4}$, \quad Yipin Zhou$^{3}$, \quad Linyan Huang$^{5}$, \\ Haoqian Wang$^{1}$\thanks{Corresponding authors.}, \quad SerNam Lim$^{3}$, \quad Hengshuang Zhao$^{2\dag}$ \\
$^{1}$Tsinghua University \ $^{2}$The University of Hong Kong \ $^{3}$ Meta AI \ $^{4}$ MIT \ $^{5}$ Xiamen University \\
{\tt\small \{lzl20, zhang-cr22\}@mails.tsinghua.edu.cn wanghaoqian@tsinghua.edu.cn hszhao@cs.hku.hk} 
}

\maketitle
\ificcvfinal\thispagestyle{empty}\fi

\begin{abstract}

In recent years, transformer-based detectors have demonstrated remarkable performance in 2D visual perception tasks. However, their performance in multi-view 3D object detection remains inferior to the state-of-the-art (SOTA) of convolutional neural network based detectors. In this work, we investigate this issue from the perspective of bird's-eye-view (BEV) feature generation. Specifically, we examine the BEV feature generation method employed by the transformer-based SOTA, BEVFormer, and identify its two limitations: (i) it only generates attention weights from BEV, which precludes the use of lidar points for supervision, and (ii) it aggregates camera view features to the BEV through deformable sampling, which only selects a small subset of features and fails to exploit all information. To overcome these limitations, we propose a novel BEV feature generation method, dual-view attention, which generates attention weights from both the BEV and camera view. This method encodes all camera features into the BEV feature. By combining dual-view attention with the BEVFormer architecture, we build a new detector named VoxelFormer. Extensive experiments are conducted on the nuScenes benchmark to verify the superiority of dual-view attention and VoxelForer. We observe that even only adopting 3 encoders and 1 historical frame during training, VoxelFormer still outperforms BEVFormer significantly. When trained in the same setting, VoxelFormer can surpass BEVFormer by 4.9\% NDS point. Code is available at: \url{https://github.com/Lizhuoling/VoxelFormer-public.git}.

\end{abstract}

\section{Introduction}
\label{Sec: Introduction}

As a fundamental 3D visual perception task, multi-view 3D object detection has grabbed broad attention from both the academic and industrial communities due to its numerous applications in areas such as autonomous driving and robotic navigation \cite{brohan2022rt,geiger2012we,yu2022generalizing}. In recent literature, many methods follow the BEV detection paradigm, which first produces a unified representation in the bird’s-eye-view (BEV) space, referred to as the BEV feature, and subsequently generates detection results based on this feature \cite{philion2020lift,wang2022detr3d}. The increasing popularity of this paradigm is partly attributed to its exceptional performance and ability to support multi-sensor input \cite{liu2022bevfusion}. 

\begin{figure}[t]
    \vspace{-0.2cm}
    \centering
    \includegraphics[scale=0.7]{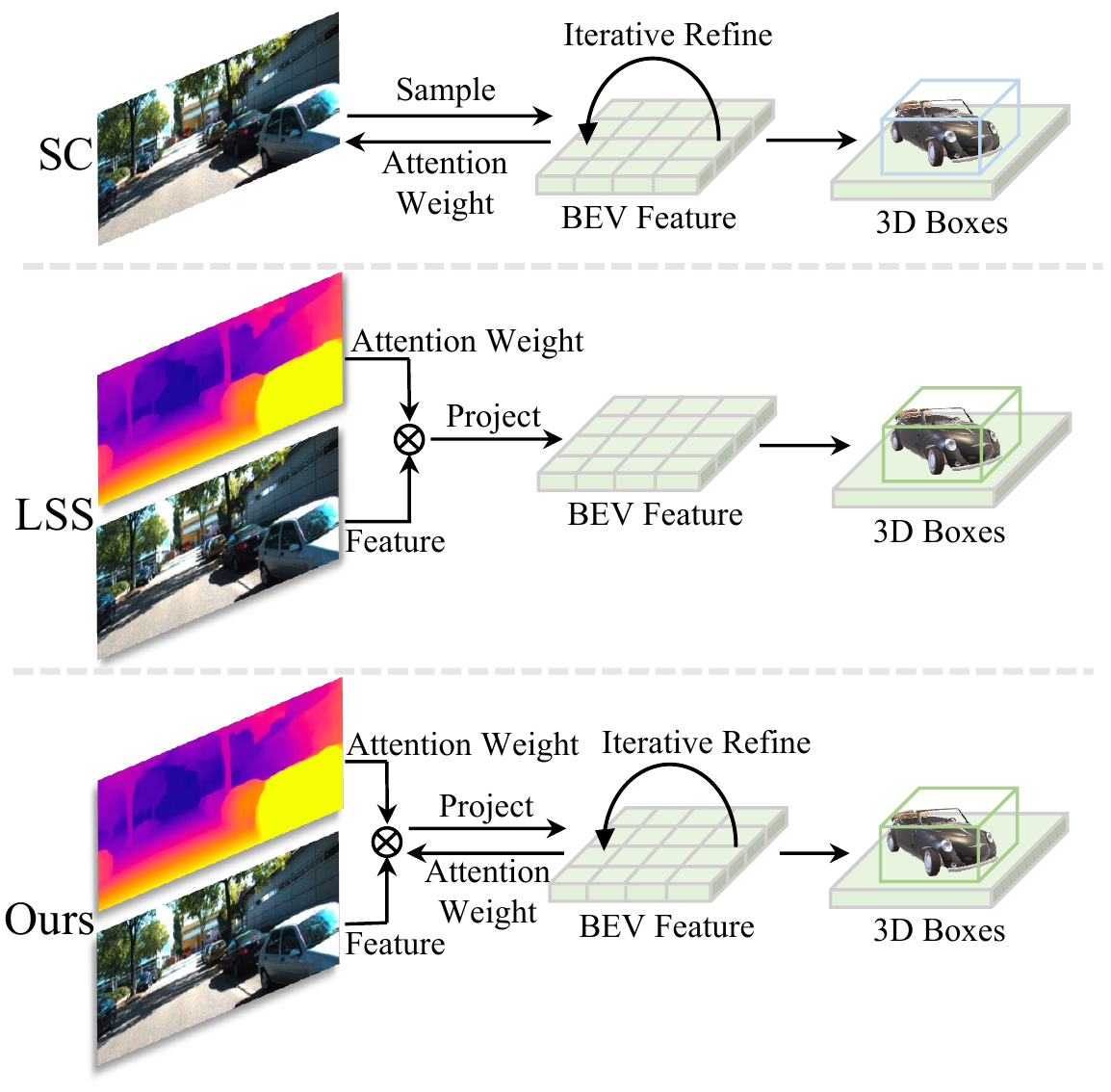}
    \caption{Comparison among three producing BEV feature methods, i.e., SC, LSS, and the proposed dual-view attention. SC predicts attention weight from the BEV while LSS generates attention weight from the camera view (the depth map). By contrast, dual-view attention produces dual attention weights from both the BEV and camera views.} \label{Fig: Comparison among producing BEV feature methods}
    \vspace{-0.4cm}
\end{figure}

There are various ways of implementing the BEV detection paradigm \cite{zhang2022beverse}, and many recent works design detectors based on Transformer \cite{carion2020end}, because Transformer-based detectors have presented promising performance in 2D visual perception tasks like 2D object detection \cite{zhang2022dino}. Nevertheless, in multi-view 3D object detection, the Transformer-based detectors still significantly lag behind the state-of-the-art (SOTA) of convolutional neural network (CNN) based counterparts, such as BEVDepth \cite{li2022bevdepth}. In this work, we hope to shed some light on this performance gap.

Since whether to produce BEV feature is the major discrepancy between multi-view 3D object detection and other 2D visual perception tasks, we delve into the aforementioned gap from the perspective of BEV feature generation. Specifically, we compare the BEV feature generation method of BEVFormer \cite{li2022bevformer} (the SOTA of Transformer-based detectors), namely spatial cross-attention (SC), with LSS \cite{philion2020lift} (the commonly adopted BEV generation method in CNN-based detectors). Through thorough analysis, we identify two significant differences, which are also illustrated in Fig.~\ref{Fig: Comparison among producing BEV feature methods}:

$\bullet$$\ $ SC produces attention weights from the BEV while LSS derives the weights (the predicted depth) from the camera view.

$\bullet$$\ $ SC obtains semantic features from the camera view through deformable sampling, which only explores a limited part of the features. By contrast, LSS directly encodes all camera view features into the BEV feature.

The first difference leads to the difficulty of employing lidar points to supervise the attention weight generation step in SC because the weight is produced in BEV, while the lidar points are typically projected to the camera view to serve as depth cues \cite{li2022delving}. Hence, detectors like BEVFormer cannot obtain depth knowledge from lidar points and are solely dependent on sparse 3D box annotations. 

Due to the second aforementioned difference, BEVFormer has to sample features from multiple Feature Pyramid Network \cite{lin2017feature} (FPN) levels based on deformable attention to increase the sample point number. This process is computationally intensive,  and SC still behaves worse in comparison to LSS, which utilizes all camera-view features.

Given the insights presented above, we propose a novel BEV feature generation method named dual-view attention, as shown in Fig.~\ref{Fig: Comparison among producing BEV feature methods}. In this method, we project all camera view features into 3D voxels and produce attention weights from both the BEV and camera view to generate BEV feature. In this way, dual-view attention allows for direct supervision from lidar points and encodes all information into the BEV feature. Additionally, as dual-view attention also predicts attention weights from BEV, it can iteratively refine the BEV feature like SC, which enables the dual-view attention to be seamlessly integrated into Transformer encoders. In Section~\ref{SubSec: Study on Producing BEV Feature Methods}, we integrate dual-view attention, SC, and LSS into the same detection architecture to compare their performances fairly. The results indicate that the developed dual-view attention surpasses the other two compared methods, SC and LSS, significantly.

Combining dual-view attention with the BEVFormer architecture, we build a new multi-view 3D object detector, namely VoxelFormer. The detector utilizes only a single level of FPN representation to derive the BEV feature. Due to the limited computing resource, we do not use the iterative historical fusion strategy in BEVFormer, which improves the result of BEVFormer on the nuScenes test set by 2.1\% NDS point but doubles the training time. Besides, only 3 encoders are adopted. We evaluate VoxelFormer using the nuScenes benchmark \cite{caesar2020nuscenes}. Trained for 24 epochs without CBGS \cite{zhu2019class}, VoxelFormer still achieves 57.4\% NDS point in the test set and outperforms BEVFormer by 0.5\% point. Additionally, as reported in Section~\ref{SubSec: Fair Comparison with BEVFormer}, when BEVFormer and VoxelFormer are compared fairly (trained in the same setting), VoxelFormer can surpass BEVFormer by 4.9\% NDS point.

\section{Related Work}
\label{Sec: Related Work}

\noindent \textbf{Camera-based 3D object detection.} Camera-based 3D object detection has been studied for years due to the expensive cost of deploying depth-aware sensors \cite{chen2016monocular}, e.g., lidar and radar. Early methods usually generate the results of every camera separately \cite{li2022diversity}. Following the design of classic 2D object detectors like Faster RCNN \cite{ren2015faster}, CenterNet \cite{zhou2019objects}, and DETR \cite{carion2020end}, most early 3D object detectors represent targets as anchors \cite{brazil2019m3d}, 2D centers \cite{liu2020smoke}, or queries \cite{wang2022detr3d}. Later, some methods represent targets as their projected 3D centers, but this strategy suffers from the object truncation problem seriously \cite{zhang2021objects}. Thanks to the emergence of the BEV detection paradigm, recent detectors first combine the features from multiple cameras into a unified representation and then directly generate 3D box results in BEV space \cite{philion2020lift}. This model design naturally avoids the aforementioned object truncation problem and presents stronger performance.

\vspace{1mm}
\noindent \textbf{BEV feature generation methods.} Existing strategies of producing the BEV feature can be categorized into two classes \cite{qian20223d}, i.e., camera-view attention and BEV attention. Both them need to predict a tensor named attention weight to transform the features from camera views as the BEV feature.  Among them, the camera-view attention based methods directly project all features from the camera view to the BEV space and predict depth maps as attention weights \cite{reading2021categorical}. These models are typically designed using CNN \cite{li2022bevdepth}. Differently, the BEV attention based methods are usually implemented based on Transformer \cite{wang2022detr3d}. They first initialize some queries in the BEV space and then use these queries to sample features from the camera view sparsely \cite{jiang2022polarformer}. The attention weights are generated from the BEV space \cite{li2022bevformer}. Comprehensively, previous methods for generating BEV features either predict attention weights from the BEV or camera view solely. The proposed dual-view attention method is the first to generate attention weights from both the BEV and camera view.

\vspace{1mm}
\noindent \textbf{Transformer.} The Transformer architecture, initially developed for natural language processing \cite{vaswani2017attention}, has been successfully adapted for use in computer vision tasks such as 2D object detection through the implementation of the DETR model \cite{carion2020end}. Unlike CNN-based models, which consist of numerous sliding convolution kernels, the Transformer utilizes the attention operation as its fundamental component, resulting in a larger receptive field \cite{ding2022scaling}. Several variants of DETR have been proposed for various computer vision applications and have achieved SOTA performance in many tasks \cite{liu2022dab,liu2021swin,zeng2022motr}. However, the performance of Transformer-based models in multi-view 3D object detection remains limited \cite{liu2022petr}. We attribute this issue mainly to two reasons: (i) The computational burden of the attention operation in the Transformer is heavy, and this issue becomes particularly pronounced in the context of 3D object detection, which requires the processing of 3D voxels instead of 2D pixels. Therefore, the Transformer-based model structure has to keep simple to save computing resources. (ii) The depth estimation part in camera-based 3D object detection is an ill-posed task \cite{li2022delving}, which further complicates the optimization of Transformer-based detectors. 

\begin{figure*}[ht]
    \vspace{-0.2cm}
    \centering
    \includegraphics[scale=0.54]{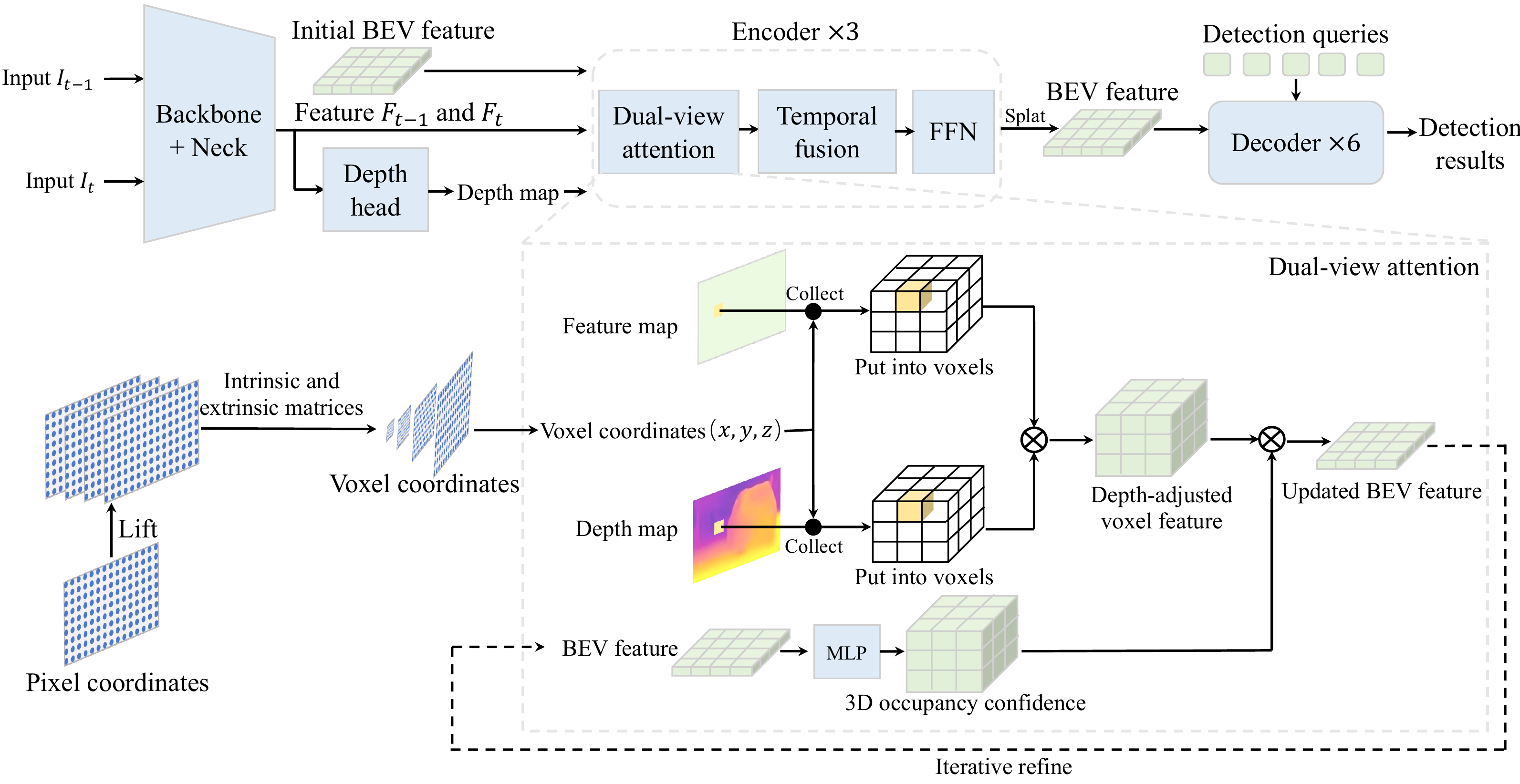}
    \caption{Overall framework of VoxelFormer.} \label{Fig: pipeline}
    \vspace{-0.4cm}
\end{figure*}

\section{Method}
\label{Sec: Method}

\subsection{Task Definition}
\label{SubSec: Task Definition}

\noindent \textbf{Multi-view 3D object detection.} In multi-view 3D object detection, there is a vehicle $V$ for perceiving the surrounding environment. $N$ monocular cameras facing towards different directions are deployed on this vehicle to capture $N$ separate images, which are the input to the 3D object detector. Notably, besides the current captured images, a limit timestamps of historical data is also available for the detector. In this work, if without special statement, only 1 historical frame is used and the training gradient is not saved for this historical frame. The lidar points collected by the lidar installed upon the vehicle $V$ can be used to assist the training process but not available during the testing phase. Given the input data, the detector needs to estimate the category, 3D center location, dimension, orientation, velocity, and attribute (such as parking or moving) of every target.

In order to explain our method clearly, we need to define several coordinate systems. The pixel coordinate system of input images is denoted as $\Psi_{p}$ and the 3D space coordinate system in the camera view is represented as $\Psi_{c}$. The points in $\Psi_{p}$ and $\Psi_{c}$ are associated with each other by the intrinsic matrix $K$. Besides, $\Psi_{c}$ is related with the vehicle ego coordinate system $\Psi_{e}$ by the extrinsic $E_{e}^{c}$, and $\Psi_{e}$ is correlated with the world coordinate system $\Psi_{w}$ by the extrinsic $E_{w}^{e}$. In this work, $\Psi_{w}$ is used for aligning the features from various timestamps, and the unified BEV feature of conducting detection is generated in $\Psi_{e}$.

\vspace{1mm}
\noindent \textbf{BEV feature generation.} Given multiple images from different cameras and timestamps as input, we first extract their features in $\Psi_{p}$. Then, we need to transform them to $\Psi_{e}$ and produce the unified BEV feature of shape ($H_{B}$, $W_{B}$, $C$), where $H_{B}$, $W_{B}$, and  $C$ are the height, width, and channel number of the BEV feature, respectively.

\subsection{VoxelFormer}
\label{SubSec: VoxelFormer}

The overall process of VoxelFormer is illustrated as Fig.~\ref{Fig: pipeline}. As shown, for the timestamp $t$, we take the current sample $I_{t} \in \mathbb{R}^{H \times W \times 3}$ and a historical sample $I_{t-1} \in \mathbb{R}^{H \times W \times 3}$ as input, where $H$ and $W$ denote the height and width of input images. $I_{t-1}$ and $I_{t}$ are first processed by the shared backbone (e.g., ResNet101 \cite{he2016deep}) and a neck (such as the Second FPN\cite{yan2018second}) to produce 2$\times N$ feature maps (the $F_{t}$ and $F_{t-1}$ in Fig.~\ref{Fig: pipeline}) corresponding to the 2$\times N$ input images. The neck takes the features with the resolutions of $\frac{H}{2^{i+1}} \times \frac{W}{2^{i+1}} \times C$ ($i=1,2,3,4$) from the backbone as input, and fuses them as a single map with the resolution of $\frac{H}{16} \times \frac{W}{16} \times C$ through upsampling and downsampling operations. Afterwards, given $F_{t-1}$ and $F_{t}$ as input, a depth estimation head is built to predict the depth of input images and obtain the depth maps $D_{t-1}$ and $D_{t}$. The feature maps ($F_{t-1}$ and $F_{t}$) and depth maps ($D_{t-1}$ and $D_{t}$) are used as input to the following encoders.

Besides the aforementioned feature maps and depth maps, another two inputs to the encoders are the initial BEV feature and voxel coordinates. The initial BEV feature $Q_{b} \in \mathbb{R}^{X_{b} \times Y_{b} \times L}$ is a randomly initialized learnable embedding consisting of $X_{b} \times Y_{b}$ vectors, and $L$ represents the length of vectors. The voxel coordinates $C_{v}$ are generated through transforming the pixel coordinates $C_{p}$ of different cameras in $\Psi_{p}$ from various timestamps to the vehicle ego coordinate system $\Psi_{e}$ at the timestamp $t$. The detailed process of generating $C_{v}$ is explained in Section \ref{SubSec: Dual-view Attention}.

Taking the image feature maps, depth maps, initial BEV feature, and voxel coordinates as input, 3 encoders are built for producing the unified BEV feature. Every encoder consists of three modules, i.e., dual-view attention, temporal fusion, and FFN. The dual-view attention module is the primary contribution of this work. It transforms the extracted 2D image features from camera views into the bev representation through generating attention weights from both the BEV and camera view. We explain the details of dual-view attention carefully in Section \ref{SubSec: Dual-view Attention}. The temporal fusion module is the same as the temporal self-attention in BEVFormer \cite{li2022bevformer}. It combines the bev features of the current and historical timestamps through conducting self-attention. The only difference is we just adopt 1 historical frame while BEVFormer totally employs 4 frames. Only using 1 historical frame and not storing its gradient can save much computing memory and training time, and our detector still outperforms BEVFormer.  FFN is a common module in visual Transformer \cite{carion2020end}, so we do not elaborate on it.

After the 3 encoders, we employ 900 detection queries and 6 decoders to decode information from the BEV feature. The implementation of the decoder follows Deformable DETR \cite{zhu2020deformable}. After the decoders, the detection queries are decoded as the concerned attributes needed for generating detetcion results (such as 3D center locations and dimensions) by a FFN module.

\subsection{Dual-view Attention}
\label{SubSec: Dual-view Attention}

The description in this part comprises three parts. In the first part, we introduce how the voxel coordinates are generated. In the second part, the process of producing the depth-adjusted voxel feature shown in Fig.~\ref{Fig: pipeline} is elaborated. Then, in the third part, how the 3D occupancy confidence and updated BEV feature are obtained are described.

\noindent \textbf{Voxel coordinates.} In this work, we divide the 3D space in the vehicle ego coordinate system $\Psi_{e}$ into uniform voxels with the shape of $(X_{b}, Y_{b}, Z)$. A zero tensor $W_{3d}$ with the shape of $(X_{b}, Y_{b}, Z, L)$ is initialized to represent these voxels. Denoting the vector with the index $(i, j, k)$ in $W_{3d}$ as $w_{3d}^{(i, j, k)}$, $w_{3d}^{(i, j, k)}$ corresponds to the $(i, j, k)$ voxel in the 3D space. In dual-view attention, the voxel coordinates $C_{v}$ contain the information that the feature vectors in feature maps ($F_{t-1}$ and $F_{t}$) and depth maps ($D_{t-1}$ and $D_{t}$) should be projected into which voxels represented in $W_{3d}$. In the following, we describe how to obtain $C_{v}$.

\begin{figure}[htbp]
    \vspace{-0.2cm}
    \centering
    \includegraphics[scale=0.4]{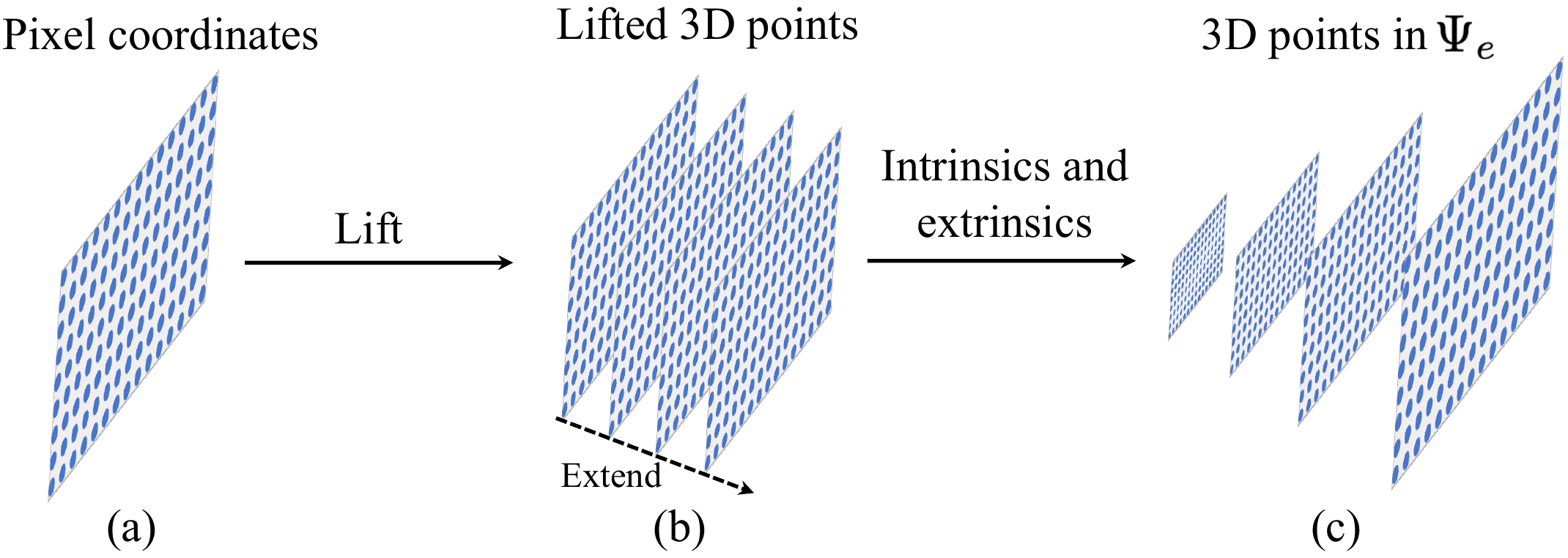}
    \caption{The procedures of lifting 2D pixels to 3D points in the vehicle ego coordinate system $\Psi_{e}$.} \label{Fig: bev_coors}
    \vspace{-0.4cm}
\end{figure}

Denoting the pixel coordinates of the feature map $F_{t}$ as $C_{p} \in \mathbb{R}^{N \times H \times W \times 2}$. We lift these pixels as 3D points by adding a depth dimension to $C_{p}$, the process of which is illustrated in Fig. \ref{Fig: bev_coors} (a) $\sim$ (b). Mathematically, it can be formulated as:
\begin{align}
C_{3d} = C_{p} \otimes C_{D},   \label{Eq1}
\end{align}
where $C_{D} = \{d_{i}\}_{i=1}^{D}$ is a vector consisting of candidate depth values, and $C_{3d} \in \mathbb{R}^{N \times H \times W \times D \times 3}$ denotes the coordinates of lifted 3D points. Then, we transform $C_{3d}$ to the vehicle ego coordinate system $\Psi_{e}$ at timestamp $t$ using camera intrinsics $K$ and extrinsics $E^{c}_{e}$, as shown in Fig. \ref{Fig: bev_coors} (b) $\sim$ (c). Specifically, for any 3D point in $C_{3d}$ with the coordinate of $(u_{3d}, v_{3d}, d_{3d})$, it is converted to the corresponding 3D point $p_{e}$ in $\Psi_{e}$ by:
\begin{align}
p_{e} = E_{e}^{c} K [u_{3d} \cdot d_{3d}, v_{3d} \cdot d_{3d}, d_{3d}]^{T}. \label{Eq2}
\end{align}
All the 3D points obtained by using Eq.~(\ref{Eq2}) compose the voxel coordinates $C_{v}$. Notably, the pixel coordinates of $F_{t-1}$ are also transformed to $\Psi_{e}$ at timestamp $t$. To this end, we first change them to the world coordinate system $\Psi_{w}$, and then to $\Psi_{e}$ at timestamp $t$.

\begin{figure}[htbp]
    \vspace{-0.2cm}
    \centering
    \includegraphics[scale=0.7]{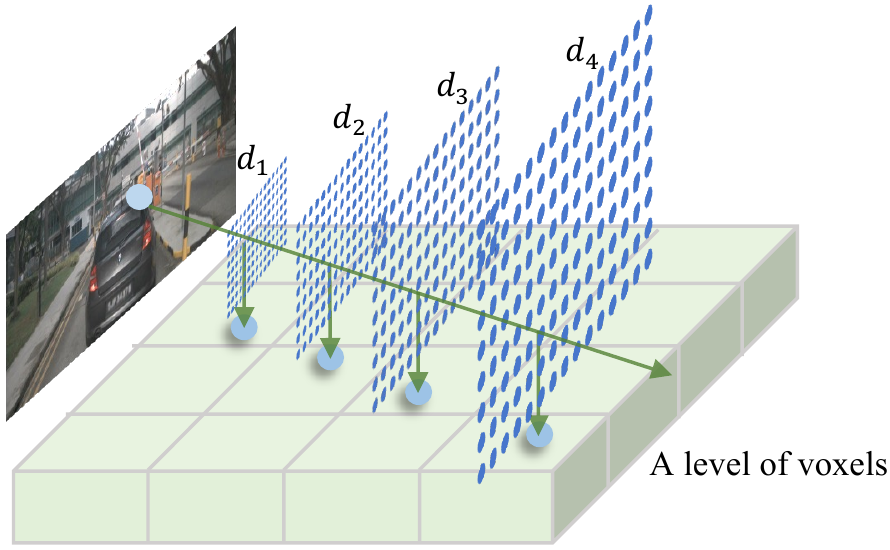}
    \caption{The process of putting image features into the voxels according to the voxel coordinates. For the simplicity of visualization, we only show a level of the voxels (the voxels of the same height).} \label{Fig: ray}
    \vspace{-0.4cm}
\end{figure}

\vspace{1mm}
\noindent \textbf{Depth-adjusted voxel feature.} Based on $C_{v}$, we know every pixel of feature vectors in feature maps ($F_{t-1}$ and $F_{t}$) and depth maps ($D_{t-1}$ and $D_{t}$) correspond to which locations in $\Psi_{e}$. Thus, we can put these feature vectors into $W_{3d}$ as shown in Fig.~\ref{Fig: ray}. 

Through putting the feature vectors in $F_{t-1}$ and $F_{t}$ into $W_{3d}$ based on $C_{v}$, we obtain the feature representation $W_{f}$. By putting $D_{t-1}$ and $D_{t}$ in $W_{3d}$, the depth representation $W_{d}$ is derived. The depth-adjusted voxel feature $W_{df}$ is produced via multiplying elements in $W_{f}$ and $W_{d}$ of the same index, which can be formulated as:
\begin{align}
w_{df}^{(i, j, k)} = w_{d}^{(i, j, k)} \cdot w_{f}^{(i, j, k)}, \label{Eq3}
\end{align}
where $w_{df}^{(i, j, k)}$, $w_{d}^{(i, j, k)}$, and $w_{f}^{(i, j, k)}$ denote the elements with the index $(i, j, k)$ in $W_{df}$, $W_{d}$, and $W_{f}$, respectively. The essence of Eq.~(\ref{Eq3}) is using the predicted depth maps as attention weights to adjust the information in the feature maps $F_{t-1}$ and $F_{t}$.

\vspace{1mm}
\noindent \textbf{3D occupancy confidence.} The process in Eq.~(\ref{Eq3}) refines the BEV feature based on the depth predicted from the camera view. Our experiments reveal that only using this producing BEV feature method does not bring not quite promising performance. We argue that this is because only adopting predicted depth maps as attention weight prevents the BEV feature generation process from being implemented in an iterative fashion, which means using the BEV feature to further refine itself. To realize this iterative fashion, it is natural to also produce attention weight based on the BEV feature.

Specifically, we predict the 3D occupancy confidence $P_{o} \in \mathbb{R}^{X_{b} \times Y_{b} \times Z}$ based on the input BEV feature $Q_{b} \in \mathbb{R}^{X_{b} \times Y_{b} \times L}$, which is implemented with a MLP layer as follows:
\begin{align}
p_{o}^{(i, j)} = MLP(q_{b}^{(i, j)}), \label{Eq4}
\end{align}
where $p_{o}^{(i, j)}$ and $q_{b}^{(i, j)}$ are the $(i, j)$ elements in $P_{o}$ and $Q_{b}$, respectively. $MLP(\cdot)$ denotes the MLP layer. 

We can observe that the 3D occupancy confidence $P_{o}$ has the same shape as the 3D space in $\Psi_{e}$, which implies that each element in $P_{o}$ corresponds to a voxel in the 3D space. In fact, every element of $P_{o}$ indicates the predicted probability that the corresponding voxel is occupied by a target. As illustrated in Fig.~\ref{Fig: pipeline}, after obtaining $P_{o}$, we further multiply it with the depth-adjusted voxel feature $W_{bf}$ and then splat the result to generate the updated BEV feature $\widetilde{Q}_{b} \in \mathbb{R}^{X_{b} \times Y_{b} \times L}$. Mathematically, it is formulated as:
\begin{align}
\widetilde{q}_{b}^{(i,j)} = \sum\limits_{k=1}^{Z} w_{df}^{(i,j,k)} \cdot p_{o}^{(i,j,k)}, 
\end{align}
where $\widetilde{q}_{b}^{(i,j)}$ denotes the $(i, j)$ element of $\widetilde{Q}_{b}$. In this way, the BEV feature is updated in the current encoder. Through repeating the aforementioned procedures in various encoders, the dual-view attention module updates the BEV feature in an iterative fashion.

\subsection{Implementation Details}
\label{SubSec: Implementation Details}

After the decoders, the detection queries are decoded as desired results by a FFN head, which includes the classification branch and regression branch. The classification branch is for estimating the category of every target and the regression branch is to regress the 3D center location, dimension, orientation, and velocity information. The attributes of targets are obtained based on pre-defined rules following DETR3D \cite{wang2022detr3d}. The depth maps obtained from lidar points are employed to train the depth head shown in Fig.~\ref{Fig: pipeline}. During the training phase, the historical frame is sampled from the past 3$\sim$27 frames (including the unlabeled frames) randomly. For the validation and testing phases, it is the $15_{\rm th}$ frame before the current frame. The model is trained for 24 epochs without CBGS, and the experiments are mainly conducted on 4 RTX3090 GPUs. 

\section{Experiment}
\label{Sec: Experiment}

\noindent \textbf{Dataset.} All the experiments are conducted using the nuScenes dataset \cite{caesar2020nuscenes}.  In this dataset, 1000 scenes of images are collected, and each scene corresponds to roughly 20s of images. The images are captured at 12 HZ by 6 cameras and annotated at 2 HZ. Besides the cameras, there are 1 lidar and 5 radars for perceiving the depth information of the surrounding environments. The scenes are officially divided into 700, 150, and 150 scenes for training, validation, and testing, respectively.

\vspace{1mm}
\noindent \textbf{Evaluation metrics.} The official metrics of nuScenes include the nuScenes Detection Score (NDS), mean Average Precision (mAP), mean Average Translation Error (mATE), mean Average Scale Error (mASE), mean Average Orientation Error (mAOE), mean Average Velocity Error (mAVE), and mean Average Attribute Error (mAAE). Among them, the NDS score reflects the overall detection performance and is the most important metric. The mAP and mATE scores reveal the localization precisions of detectors. The mASE, mAOE, mAVE, and mAAE scores indicate the performance on estimating dimension, orientation, velocity, and attribute, respectively. 

\begin{table*}[tbp] 
    \centering
    \vspace{-0.2cm}
    \resizebox{162mm}{28mm}{
    \begin{tabular}{c|cc|ccccccc}
    \hline
    Methods & Backbone & Image Size & mATE$\downarrow$  & mASE$\downarrow$ & mAOE$\downarrow$ & mAVE$\downarrow$ & mAAE$\downarrow$ & mAP$\uparrow$ & \textbf{NDS}$\uparrow$ \\
    \cline{1-10}
    FCOS3D \cite{wang2021fcos3d} & Res101 & $1600 \times 900$ & 0.690 & 0.249 & 0.452 & 1.434 & 0.124 & 0.358 & 0.428 \\
    PGD \cite{wang2022probabilistic} & Res101 & $1600 \times 900$ & 0.626 & 0.245 & 0.451 & 1.509 & 0.127 & 0.386 & 0.448 \\
    BEVDet \cite{huang2021bevdet} & ST-Small & $1600 \times 640$ & 0.556 & 0.239 & 0.414 & 1.010 & 0.153 & 0.398 & 0.463 \\
    DD3D \cite{park2021pseudo} & VoVNet & $1600 \times 900$ & 0.572 & 0.249 & 0.368 & 1.014 & 0.124 & 0.418 & 0.477 \\
    BEVDet4D \cite{huang2022bevdet4d} & VoVNet & $1600 \times 640$ & 0.511 & 0.241 & 0.386 & 0.301 & 0.121 & 0.451 & 0.569 \\
    BEVDepth \cite{li2022bevdepth} & VoVNet & $1600 \times 640$ & 0.445 & 0.245 & 0.378 & 0.320 & 0.126 & 0.503 & {\color{blue}0.600} \\
    \cline{1-10}
    DETR3D \cite{wang2022detr3d} & VoVNet & $1600 \times 900$ & 0.641 & 0.255 & 0.394 & 0.845 & 0.133 & 0.412 & 0.479 \\
    PETR \cite{liu2022petr} & VoVNet & $1600 \times 640$ & 0.593 & 0.249 & 0.383 & 0.808 & 0.132 & 0.441 & 0.504 \\
    BEVFormer \cite{li2022bevformer} & VoVNet & $1600 \times 900$ & 0.582 & 0.256 & 0.375 & 0.378 & 0.126 & 0.481 & 0.569 \\
    PolarFormer \cite{jiang2022polarformer} & VoVNet & $1600 \times 900$ & 0.556 & 0.256 & 0.364 & 0.439 & 0.127 & 0.493 & 0.572 \\
    \cline{1-10}
    VoxelFormer (Ours) & VoVNet & $800 \times 320$ & 0.562  & 0.255  & 0.386 & 0.402 & 0.122 & 0.465 & 0.560 \\
    VoxelFormer (Ours) & VoVNet & $1600 \times 640$ & 0.538 & 0.257 & 0.366 & 0.390 & 0.135 & 0.486 & {\color{red}0.574} \\
    \hline
    \end{tabular}}
    \caption{Comparison with previous SOTA counterparts on the nuScenes test set. The $1_{\rm st} \sim 6_{\rm th}$ rows of results are CNN-based methods, and the $7_{\rm th} \sim 10_{\rm th}$ rows are Transformer-based methods. Our method is also Transformer-based. The NDS score is the primary metric. The CNN-based SOTA is marked in {\color{blue}blue} and the best performance among the listed Transformer-based methods is highlighted in {\color{red}red}. Compared with the baseline method BEVFormer, VoxelFormer only employs 1 historical sample in the training phase and adopts 3 encoders.} \label{Table: Comparison on the test set}
    \vspace{-0.2cm}
\end{table*}

\begin{table*}[tbp] 
    \centering
    \resizebox{162mm}{24mm}{
    \begin{tabular}{c|cc|ccccccc}
    \hline
    Methods & Backbone & Image Size & mATE$\downarrow$  & mASE$\downarrow$ & mAOE$\downarrow$ & mAVE$\downarrow$ & mAAE$\downarrow$ & mAP$\uparrow$ & \textbf{NDS}$\uparrow$ \\
    \cline{1-10}
    FCOS3D \cite{wang2021fcos3d} & Res101 & $1600 \times 900$ & 0.806 & 0.268 & 0.511 & 1.315 & 0.170 & 0.295 & 0.372 \\
    DETR3D \cite{wang2022detr3d} & Res101 & $1600 \times 900$ & 0.860 & 0.278 & 0.437 & 0.967 & 0.235 & 0.303 & 0.374 \\
    BEVDet \cite{huang2021bevdet} & Res101 & $1056 \times 384$ & 0.704 & 0.273 & 0.531 & 0.940 & 0.250 & 0.317 & 0.389 \\
    PETR \cite{liu2022petr} & Res101 & $1408 \times 512$ & 0.710 & 0.270 & 0.490 & 0.885 & 0.224 & 0.357 & 0.421 \\
    BEVFormer-S \cite{li2022bevformer} & Res101 & $1600 \times 900$ & 0.725 & 0.272 & 0.391 & 0.802 & 0.200 & 0.375 & 0.448 \\
    VoxelFormer-S & Res101 & $1600 \times 640$ & 0.694 & 0.267 & 0.441 & 0.725 & 0.205 & 0.406 & \textbf{0.470} \\
    \cline{1-10}
    BEVFormer \cite{li2022bevformer} & Res101 & $1600 \times 900$ & 0.673 & 0.274 & 0.372 & 0.394 & 0.198 & 0.416 & 0.517 \\
    PolarFormer \cite{jiang2022polarformer} & Res101 & $1600 \times 900$ & 0.648 & 0.270 & 0.348 & 0.409 & 0.201 & 0.432 & 0.528 \\
    VoxelFormer-res & Res101 & $1600 \times 640$ & 0.663 & 0.271 & 0.371 & 0.342 & 0.182 & 0.426 & 0.530\\
    VoxelFormer-vov & VoVNet & $1600 \times 640$ & 0.647 & 0.269 & 0.364 & 0.355 & 0.195 & 0.451 & \textbf{0.542} \\
    \hline
    \end{tabular}}
    \caption{Comparison with previous methods on the nuScenes validation set. The $1_{\rm st}$ to $6_{\rm th}$ rows of results correspond to detectors without using historical information, and the $7_{\rm th}$ to $10_{\rm th}$ rows are detectors with historical information. The best results are marked in \textbf{bold}.} \label{Table: Comparison on the validation set}
    \vspace{-0.4cm}
\end{table*}

\noindent \textbf{Experimental settings.} For the results on the testing set, we train detectors on both the training and validation sets of nuScenes. Conversely, the results on the validation set are from detectors trained on only the training set. Due to our limited computing resource, it is difficult to conduct all experiments in the full setting (the backbone is VoVNet, $X_{b}=200$, $Y_{b}=200$, $Z=8$, input image resolution is $1600 \times 640$, 1 historical frame). Therefore, except the experiments in Section~\ref{SubSec: Comparison on with Previous Methods}, the remaining experiments are primarily conducted in the half setting (the backbone is ResNet50, $X_{b}=128$, $Y_{b}=128$, $Z=8$, input image resolution is $800 \times 320$, no historical frame). The optimizer and hyper-parameter settings all follow BEVFormer \cite{li2022bevformer}.

\subsection{Comparison on with Previous Methods}
\label{SubSec: Comparison on with Previous Methods}

\noindent \textbf{nuScenes test set.} We compare the proposed detector, VoxelFormer, with previous methods on the nuScenes test set in Table~\ref{Table: Comparison on the test set}. It can be observed that VoxelFormer surpasses all Transformer-based methods, especially the baseline method BEVFormer, significantly. Notably, due to the limited computing resource, we do not use the iterative temporal fusion training strategy (3 historical samples for training) in BEVFormer, which brings 2.7\% NDS improvement. Besides, we only employ 3 encoders. The results in Section \ref{SubSec: Ablation Study} suggest that the detection accuracy of VoxelFormer can be further improved if we also adopt 6 encoders. Moreover, we do not use CBGS and only employ 1 level of FPN feature to produce the BEV feature. In such a naive training setting, our method still behaves the best compared with other Transformer-based detectors, which indicates the superiority of the proposed dual-view attention module.

In addition, it can be observed from the $11_{\rm th} \sim 12_{\rm th}$ rows of results in Table~\ref{Table: Comparison on the test set}, the VoxelFormer with half input image resolution ($800 \times 320$) obtains 56.0\% NDS point, which is only 1.4\% NDS point lower than the full input image resolution setting ($1600 \times 640$). This observation further confirms the efficiency of VoxelFormer. We attribute this efficiency to the proposed dual-view attention module, which generates feature attention weights ($W_{f}$ and $P_{o}$) from both the BEV and camera view.

\vspace{1mm}
\noindent \textbf{nuScenes validation set.} Besides the nuScenes test set, we also compare VoxelFormer with previous detectors in the validation set. The results are presented in Table~\ref{Table: Comparison on the validation set}. The $1_{\rm st}$ to $6_{\rm th}$ rows of results in Table~\ref{Table: Comparison on the validation set} do not use historical frames, and the remaining rows correspond to detectors with historical information. As shown, VoxelFormer outperforms all compared methods in both two settings, i.e., without historical information and using one historical sample. The performance superiority is especially significant in the setting of without historical information, where VoxelFormer-S surpasses the baseline BEVFormer-S by 2.2\% NDS score.

\subsection{Fair Comparison with BEVFormer}
\label{SubSec: Fair Comparison with BEVFormer}

\noindent \textbf{Performance comparison.} Since BEVFormer is the baseline of VoxelFormer, comparing BEVFormer and VoxelFormer is the most direct way to demonstrate the superiority of VoxelFormer. Nevertheless, for the experiments in Section~\ref{SubSec: Comparison on with Previous Methods}, the comparison between them is exactly unfair because BEVFormer uses more encoders and historical frames. This difference conceals the advantage of VoxelFormer to some extent. In this experiment, both BEVFormer and VoxelFormer adopt 3 encoders. We compare them in both the settings using 0 and 1 historical frame. The experimental results are reported in Table~\ref{Table: fair comparison}.

\begin{table}[htbp] 
    \centering
    \vspace{-0.2cm}
    \resizebox{82mm}{9mm}{
    \begin{tabular}{c|c|ccccc}
    \hline
    Method & HFN & mATE$\downarrow$  & mASE$\downarrow$ & mAOE$\downarrow$ & mAP$\uparrow$ & \textbf{NDS}$\uparrow$ \\
    \cline{1-7}
    \multirow{2}{*}{BEVFormer} & 0 & 0.984 & 0.294 & 0.758 & 0.195 & 0.270 \\
    \cline{2-7}
    & 1 & 0.871 & 0.292 & 0.689 & 0.272 & 0.357 \\
    \cline{1-7}
    \multirow{2}{*}{VoxelFormer} & 0 & 0.844 & 0.290 & 0.723 & 0.264 & 0.319 \\
    \cline{2-7}
    & 1 & 0.860 & 0.287 & 0.615 & 0.268 & 0.386 \\
    \hline
    \end{tabular}}
    \caption{Fair comparison between BEVFormer and VoxelFormer on the detection performance (HFN: historical frame number).} \label{Table: fair comparison}
    \vspace{-0.2cm}
\end{table}

As presented in Table~\ref{Table: fair comparison}, VoxelFormer outperforms BEVFormer by large margins in both settings. Specifically, when no historical information is used, VoxelFormer surpasses BEVFormer by 4.9\% NDS point. When 1 historical sample is employed, VoxelFormer outperforms BEVFormer by 2.9\% NDS point. Since the main difference between these two methods is that VoxelFormer produces the BEV feature based on the proposed dual-view attention module, these results confirm the effectiveness of the dual-view attention. In addition, according to the aforementioned results, we can observe that the dual-view attention is especially effective when historical information is unavailable.

\noindent \textbf{Inference speed and parameter volume.} In this part, we compare the inference speeds and parameter volumes of BEVFormer and VoxelFormer using a RTX3090 GPU. Both them do not use historical information and adopt 3 encoders. The adopted backbone is ResNet-101. The results are given in Table~\ref{Table: speed and parameter}.

\begin{table}[htbp] 
    \vspace{-0.2cm}
    \centering
    \resizebox{75mm}{6.1mm}{
    \begin{tabular}{c|cc}
    \hline
    Method & Inference Speed (FPS)$\uparrow$  & Parameter Volume (k)$\downarrow$ \\
    \cline{1-3}
    BEVFormer & 7.634 & 180.93 \\
    VoxelFormer & 8.197 & 164.48 \\
    \hline
    \end{tabular}}
    \caption{Fair comparison between BEVFormer and VoxelFormer on the inference speed and parameter volume. FPS refers to frame per second.} \label{Table: speed and parameter}
    \vspace{-0.2cm}
\end{table}

According to the results in Table~\ref{Table: speed and parameter}, VoxelFormer behaves faster than BEVFormer and contains fewer parameters. Meanwhile, as shown in the results of Table~\ref{Table: fair comparison}, the performance of VoxelFormer is significantly better than BEVFormer. This observation further demonstrates the superiority of the proposed VoxelFormer.

\begin{figure*}[tbp]
    \centering
    \includegraphics[scale=0.6]{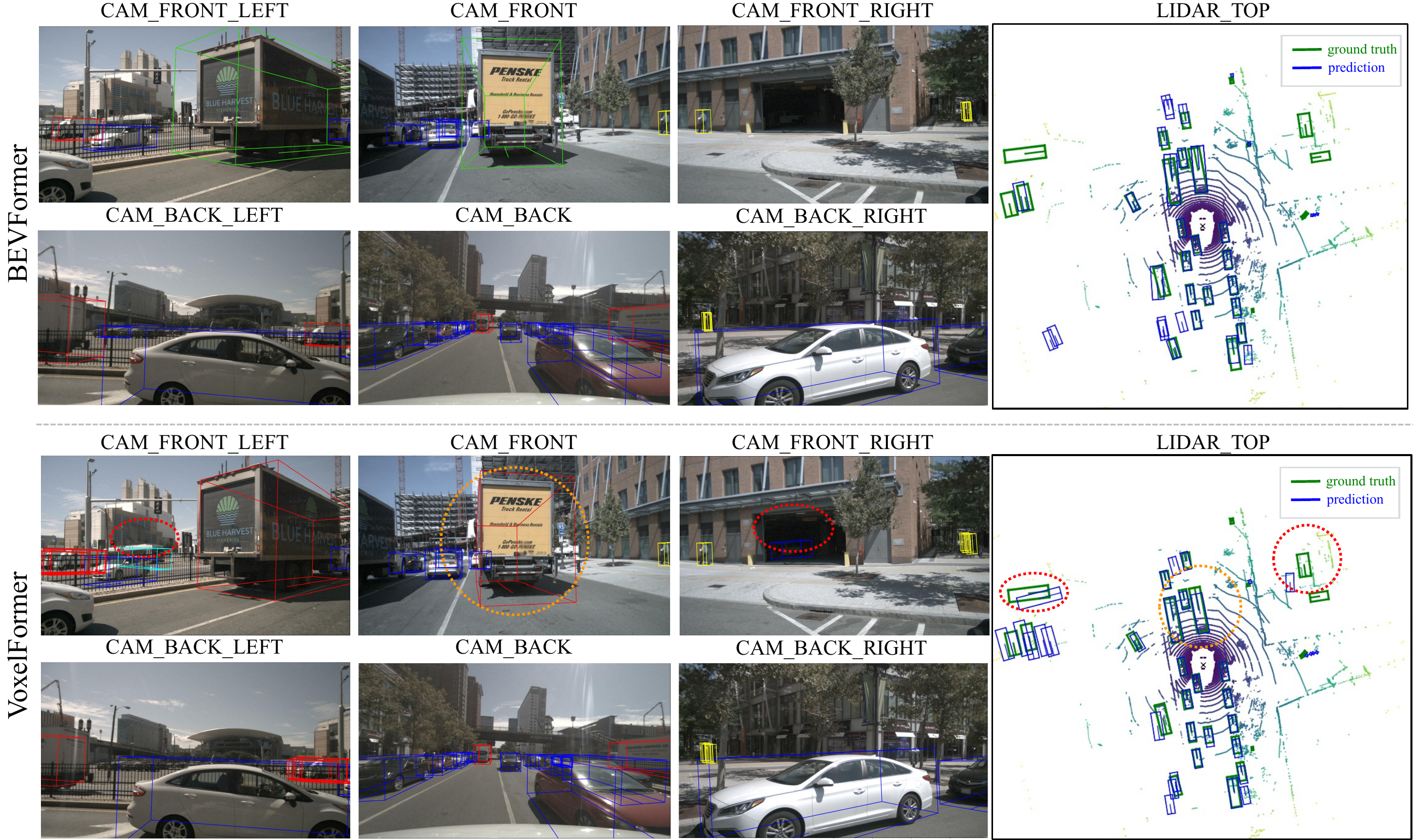}
    \caption{Visualization of the detection results produced by BEVFormer and VoxelFormer in the nuScenes validation set. The {\color{orange}orange} box corresponds to the target that is categorized into a false class by BEVFormer but classified accurately by VoxelFormer. The {\color{red}red} circles are targets that are not detected by BEVFormer but localized correctly by VoxelFormer.} \label{Fig: vis}
    \vspace{-0.4cm}
\end{figure*}

\subsection{Study on BEV Feature Generation Methods}
\label{SubSec: Study on Producing BEV Feature Methods}

The most effective way of confirming the effectiveness of the primary contribution, dual-view attention, is integrating it and its counterparts into the same detector and compare their resulted detection performances. In this way, the complex differences in implementation details of various detectors are avoided, and we can concentrate on the comparison among different producing BEV feature methods.

To the aforementioned end, we integrate dual-view attention and the two existing mainstream BEV feature generation methods, SC and LSS, into the VoxelFormer detector in the half setting. The results are reported in Table~\ref{Table: comparison among producing BEV feature methods}.

\begin{table}[htbp] 
    \centering
    \vspace{-0.2cm}
    \resizebox{82mm}{7mm}{
    \begin{tabular}{c|ccccccc}
    \hline
    Method & mATE$\downarrow$  & mASE$\downarrow$ & mAOE$\downarrow$ & mAVE$\downarrow$ & mAAE$\downarrow$ & mAP$\uparrow$ & \textbf{NDS}$\uparrow$ \\
    \cline{1-8}
    SC & 0.872 & 0.294 & 0.776 & 1.085 & 0.323 & 0.257 & 0.302 \\
    LSS & 0.848 & 0.294 & 0.735 & 1.121 & 0.350 & 0.254 & 0.305 \\
    DVA & 0.842 & 0.294 & 0.702 & 0.969 & 0.254 & 0.264 & 0.326 \\
    \hline
    \end{tabular}}
    \caption{Comparison among various producing BEV feature methods in the same detector architecture. DVA refers to the dual-view attention.} \label{Table: comparison among producing BEV feature methods}
    \vspace{-0.2cm}
\end{table}

According to the numeric results in Table~\ref{Table: comparison among producing BEV feature methods}, the proposed dual-view attention achieves the best performance considering all the metrics. Besides, comparing the $1_{\rm st}$ and $2_{\rm nd}$ rows of results in this table, we can observe that although the LSS is designed for CNN-based 3D object detectors, its performance of using a Transformer-based detector is still promising and slightly better than SC, the one specially designed for Transformer-based detectors. Specifically, LSS behaves better than SC in the metrics of reflecting localization and orientation estimation performances, i.e., mATE, mAOE, and mAP. However, SC outperforms LSS in the metrics of velocity estimation and attribute classification, mAVE and, mAAE.

\subsection{Ablation Study}
\label{SubSec: Ablation Study}

\noindent \textbf{The effect of the depth attention weight.} As introduced in Section~\ref{SubSec: Dual-view Attention}, the dual-view attention produces attention weights from both the BEV and camera view. Compared with the previous BEV feature generation method (the SC) of Transformer-based detectors, the main difference is we also use the depth attention weight predicted from the camera view, which is exactly the depth maps corresponding to input images. In this part, we analyze how the depth attention weight affects the overall performance. The results of using and without using the depth attention weight are reported in the $1_{\rm st}$ and $2_{\rm nd}$ rows of results in Table~\ref{Table: depth attention weight}.

\begin{table}[htbp] 
    \centering
    \vspace{-0.2cm}
    \resizebox{82mm}{6mm}{
    \begin{tabular}{c|ccccccc}
    \hline
    Depth & mATE$\downarrow$  & mASE$\downarrow$ & mAOE$\downarrow$ & mAVE$\downarrow$ & mAAE$\downarrow$ & mAP$\uparrow$ & \textbf{NDS}$\uparrow$ \\
    \cline{1-8}
    No & 0.899 & 0.295 & 0.727 & 0.960 & 0.265 & 0.249 & 0.310 \\
    Yes & 0.842 & 0.294 & 0.702 & 0.969 & 0.254 & 0.264 & 0.326 \\
    \hline
    \end{tabular}}
    \caption{Ablation study on whether to use predicted depth as attention weight in the dual-view attention.} \label{Table: depth attention weight}
    \vspace{-0.2cm}
\end{table}

According to Table~\ref{Table: depth attention weight}, the depth attention weight boosts the performance of VoxelFormer by 1.4\% NDS point. This observation confirms the effectiveness of the depth attention weight. We attribute this improvement to the employment of lidar points as explicit depth supervision, which alleviates the optimization difficulty of the depth head.

\vspace{1mm}
\noindent \textbf{Effect of encoder and decoder numbers.} In this part, we study how the numbers of encoders and decoders affect the evaluation scores on the nuScenes validation set. The experimental results are given in Table~\ref{Table: Encoder and decoder number}.

\begin{table}[htbp] 
    \centering
    \vspace{-0.2cm}
    \resizebox{82mm}{12mm}{
    \begin{tabular}{cc|ccccc}
    \hline
    Encoder & Decoder & mATE$\downarrow$  & mASE$\downarrow$ & mAOE$\downarrow$ & mAP$\uparrow$ & \textbf{NDS}$\uparrow$ \\
    \cline{1-7}
    1 & 6 & 0.902 & 0.291 & 0.632 & 0.219 & 0.347 \\
    3 & 6 & 0.893 & 0.286 & 0.609 & 0.225 & 0.357 \\
    6 & 1 & 0.937 & 0.315 & 0.781 & 0.203 & 0.304 \\
    6 & 3 & 0.874 & 0.296 & 0.672 & 0.246 & 0.362 \\
    6 & 6 & 0.878 & 0.290 & 0.643 & 0.250 & 0.371 \\
    \hline
    \end{tabular}}
    \caption{Ablation study on the encoder and decoder numbers using the nuScenes validation set.} \label{Table: Encoder and decoder number}
    \vspace{-0.2cm}
\end{table}

As shown, with the increase of encoder or decoder numbers, the detection precision is continuously boosted. For example, according to the $2_{\rm nd}$ and $3_{\rm rd}$ rows of results in Table~\ref{Table: Encoder and decoder number},  the NDS score in the setting of employing 6 encoders is 1.4\% point higher than the setting of only using 3 encoders. This observation suggests that the performance of VoxelFormer on the nuScenes test set can be further improved by adopting 6 encoders. However, due to limited computing resources, we only use 3 encoders. 

\subsection{Visualization}
\label{SubSec: Visualization}

We visualize the detection results of BEVFormer and VoxelFormer in Fig.~\ref{Fig: vis}. As shown, although some cases are very challenging and not accurately classified or correctly localized by BEVFormer, they are detected successfully by VoxelFormer. This result reveals the promising detection precision of the developed detector, VoxelFormer.

\section{Conclusion and Limitation}
\label{Sec: Conclusion and Limitation}

In this work, a novel BEV feature generation method, named dual-view attention, has been proposed. Compared with previous methods, it generates attention weights from both the BEV and camera view. In this way, it can use lidar points as explicit depth supervision and encode all information from cameras into the BEV feature. By combining the dual-view attention and the BEVFormer architecture, we have built a new detector, VoxelFormer. Extensive experiments have been conducted on the nuScenes benchmark to confirm the superiority of the dual-view attention and VoxelFormer. The primary limitation of this work is we do not train our models using the same setting as the newest SOTA detectors, such as using 6 encoders, CBGS, more historical frames, etc. This is because our limited computing resources cannot support such an expensive training setting.

\begin{appendix}

\renewcommand{\appendixname}{\Appendix~\Alph{section}}

\section{Appendix}

\subsection{Efficient CUDA Implementation}

Previous BEV feature generation methods produce attention weight solely from the BEV or camera view feature \cite{li2022bevdepth,li2022bevformer}. By contrast, the proposed dual-view attention derives attention weight based on both the BEV and camera view features. However, the dual-view attention presents faster inference speed and consumes relatively smaller GPU memory, because we implement it efficiently in CUDA. Since dual-view attention is not responsible for fusing information of the current sample and the historical sample, its computing process for the current and historical samples is the same. Therefore, in the following, we only explain how the current sample is processed by the dual-view attention implemented in CUDA for simplicity.

The process of dual-view attention is introduced in Section~\ref{SubSec: Dual-view Attention} of the paper. The 3D occupancy confidence $P_o \in \mathbb{R}^{X_b \times Y_b \times Z}$ is predicted by an Linear layer provided by Pytorch before input to the dual-view attention CUDA operator. The dual-view attention operator has 4 inputs, i.e., the feature map $F_t \in \mathbb{R}^{N \times H \times W \times L}$, depth map $D_t \in \mathbb{R}^{N \times H \times W \times D}$, voxel coordinates $C_{3d} \in \mathbb{R}^{N \times H \times W \times D \times 3}$, and 3D occupancy confidence $P_o$. All the hyper-parameters ($N$, $H$, $W$, $D$, $X_b$, $Y_b$, and $L$) have been defined in the paper. $F_t$, $D_t$, and $C_{3d}$ only need to be computed once and thus are generated before encoders. $P_o$ is predicted in every encoder before the dual-view attention CUDA operator.

In dual-view attention, the process described in Section 3.3 of the paper can be summarized as multiplying corresponding elements from $F_t$, $D_t$, and $P_o$ based on $C_{3d}$, and then adding the obtained feature vectors to the corresponding voxels. Specifically, $N \times H \times W \times D$ threads are initialized in a GPU. We take the thread with the index $(n, h, w, d)$ as an example to decribe the computing process. Assuming the generated voxel feature as $W_{3d} \in \mathbb{R}^{N \times X_b \times Y_b \times Z \times L}$, the computing process of the thread with the index $(n, h, w, d)$ is described in Algorithm~\ref{CUDA implementation of dual-view attention}. After the operations in CUDA, the updated BEV feature $\tilde{Q}_b \in \mathbb{R}^{N \times X_b \times Y_b \times L}$ is obtained by summing the elements of $W_{3d}$ in the height dimension.

We can observe from Algorithm~\ref{CUDA implementation of dual-view attention} that the forwarding computation process of dual-view attention is quite simple, which can be realized with only one line of code. Therefore, the inference speed of dual-view attention and VoxelFormer is promising. Besides, our cuda implementation of dual-view attention does not produce a 5D feature tensor like the voxel pooling operator in BEVDepth, which saves much GPU memory.

\begin{algorithm}[tb]
\caption{Pseudo code for a single thread in the dual-view attention CUDA operator.} \label{CUDA implementation of dual-view attention}
\begin{algorithmic}[1]
    \REQUIRE Feature map $F_t$, depth map $D_t$, voxel coordinates $C_{3d}$, 3D occupancy confidence $P_o$, voxel feature $W_{3d}$, and thread index $(n, h, w, d)$
    \STATE $(x_b, y_b, z) = C_{3d}[n, h, w, d]$
    \STATE $W_{3d}[x_b, y_b, z] \leftarrow W_{3d}[x_b, y_b, z] + F_{t}[n, h, w] \cdot D_{t}[n, h, w, d] \cdot P_{o}[x_b, y_b, z]$
    
    \ENSURE $W_{3d}$
\end{algorithmic}
\end{algorithm}

\end{appendix}

{\small
\bibliographystyle{ieee_fullname}
\bibliography{reference}
}

\end{document}